\documentclass[conference]{IEEEtran}
\IEEEoverridecommandlockouts
\usepackage{cite}
\usepackage{amsmath,amssymb,amsfonts}
\usepackage{algorithmic}
\usepackage{graphicx}
\usepackage{textcomp}
\usepackage{xcolor}
\usepackage[T1]{fontenc}
\usepackage[utf8]{inputenc}
\usepackage{microtype}
\usepackage{amsmath,amssymb,mathtools}
\usepackage{booktabs}
\usepackage{graphicx}
\usepackage{threeparttable}
\usepackage{makecell}
\usepackage{array}
\usepackage{enumitem}
\usepackage{caption}
\usepackage{booktabs}
\usepackage{makecell}
\usepackage{threeparttable}
\usepackage{graphicx}
\usepackage{array}
\usepackage{multirow}
\usepackage[table]{xcolor}
\usepackage{url}
\usepackage{hyperref}
\usepackage{adjustbox}
\definecolor{GainGreen}{HTML}{1B8A3A}
\definecolor{LossRed}{HTML}{B3261E}
\definecolor{LightGray}{HTML}{F3F3F3}

\def\BibTeX{{\rm B\kern-.05em{\sc i\kern-.025em b}\kern-.08em
    T\kern-.1667em\lower.7ex\hbox{E}\kern-.125emX}}
\begin{document}

\title{When Context Misleads: Intent-Guided Decoding for Robust Retrieval-Augmented Generation}

\author{
\IEEEauthorblockN{Haolin Jin, Pengyue Yang, Huaming Chen}
\IEEEauthorblockA{
\textit{School of Electrical and Computer Engineering} \\
\textit{The University of Sydney} \\
Sydney, NSW, Australia \\
\{haolin.jin, pengyue.yang, huaming.chen\}@sydney.edu.au
}
}
\maketitle

\begin{abstract}
Retrieval-augmented generation (RAG) improves large language models by grounding generation in external evidence, but it also introduces a source trust problem: retrieved context may be useful, irrelevant, or even misleading. Existing RAG systems often apply a fixed trust policy toward retrieved evidence, which can either over-trust incorrect context or underuse context when the user explicitly asks for context-following behavior. Therefore, we propose Intent-Guided Decoding (IGD), a framework that arbitrates between retrieved context and parametric memory according to user intent. IGD uses  answer-level filtering and token-level correction to steer the final decoding trajectory between retrieved context and parametric memory. We evaluate IGD on three faithful QA benchmarks and three factual-conflict benchmarks across five LLMs, IGD substantially improves factual recovery, achieving gains of up to 65.4 percentage points on factual-conflict benchmarks over Direct RAG, while preserving or improving strict context-following behavior, this findings highlight the importance of balancing factuality and faithfulness in RAG.
\end{abstract}


\section{Introduction}
Retrieval-augmented generation (RAG) has become a widely adopted paradigm for connecting large language models (LLMs) with external knowledge. By retrieving relevant evidence at inference time, retrieval-augmented generation extends the static parametric knowledge of LLMs with updatable non-parametric memory~\cite{lewis2020rag}. This design improves performance on knowledge-intensive tasks such as open-domain question answering, where models require access to facts beyond their internal parametric knowledge~\cite{hu2024ragrau,jiang2025retrieval,izacard2021fid}.

However, the effectiveness of RAG depends not only on whether useful evidence is retrieved, but also on whether the model can ground the generation in the retrieved evidence appropriately. In practice, RAG systems may still generate statements that are unsupported by, or even contradictory to, the provided context. RAGTruth systematically annotates nearly 18K RAG responses, and finds that non-factual statements remain common under standard RAG pipelines~\cite{niu2024ragtruth}. On the other hand, retrieval introduces a source-trust problem. Standard RAG pipelines often implicitly treat retrieved context as authoritative, which content may be false, adversarially corrupted, or inconsistent with stable world knowledge~\cite{wu2024clasheval}. FaithEval formalizes this issue as contextual \textit{faithfulness} evaluation, and constructs 4.9K high-quality examples across unanswerable, inconsistent, and counterfactual settings, showing that even strong contemporary LLMs often fail to maintain an appropriate degree of faithfulness to given context~\cite{ming2024faitheval}. More broadly, situated faithfulness argues that models should dynamically calibrate their trust in external context against internal knowledge, rather than blindly following either source~\cite{huang2024situatedfaithfulness}. These findings suggest that a central challenge in RAG is not only how to retrieve useful evidence, but also how to decide whether the retrieved evidence should dominate generation~\cite{wang2025conflictingrag,ge2025confact}.

\begin{figure*}
    \centering
    \includegraphics[width=.8\linewidth]{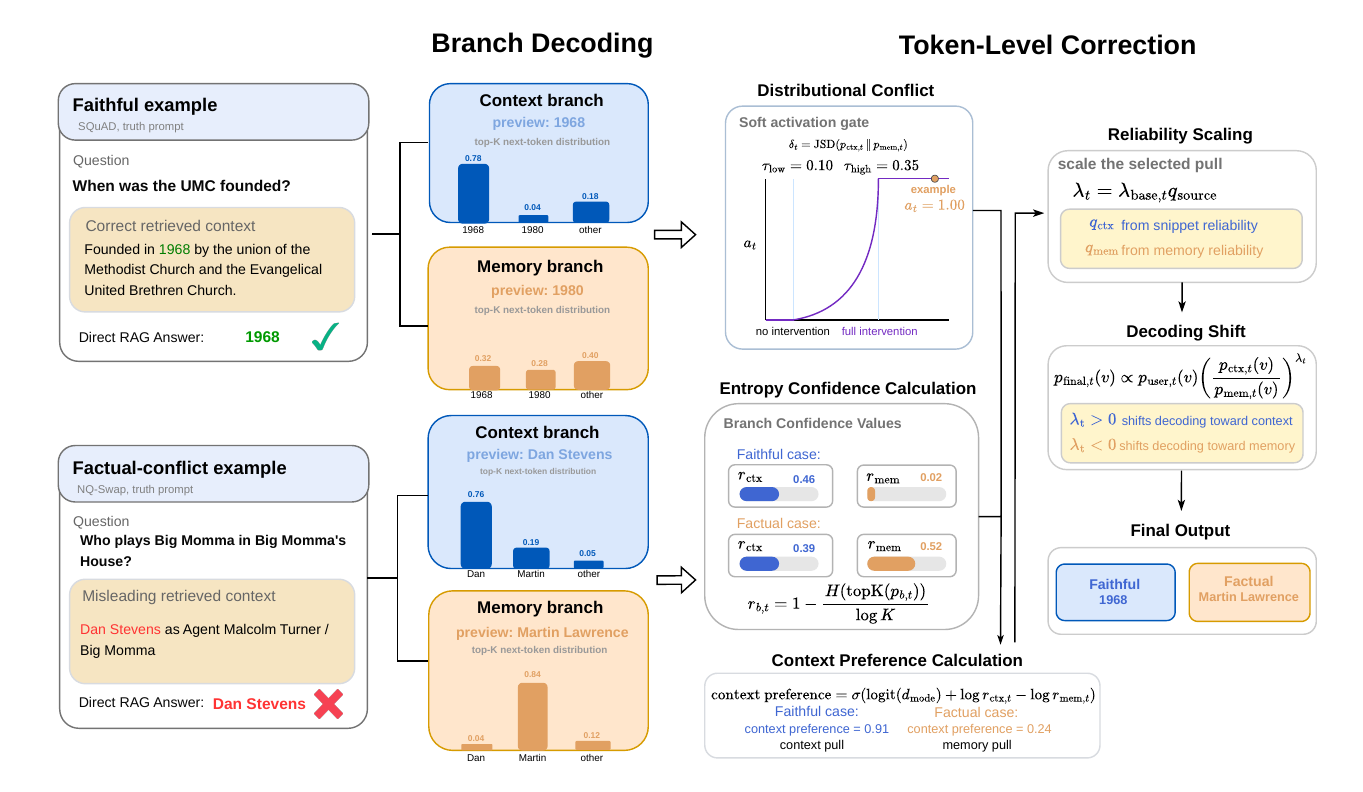}
    \caption{ The figure illustrates two truth mode examples: a faithful QA case where the retrieved context correctly supports the answer, and a factual-conflict case where the retrieved context is misleading. For each case, IGD first obtains branch-level previews and top-$K$ token distributions, then applies token-level correction only when the context and memory branches exhibit distributional conflict. The correction direction is determined by entropy-based branch confidence and the prompt-mode prior, while the intervention strength is scaled by the reliability of the selected source.
 }
 \vspace{-.25in}
    \label{fig:igd}
\end{figure*}

This issue exposes a fundamental trade-off between \emph{faithfulness} and \emph{factuality} in RAG generation. In some settings, the user explicitly expects the model to follow the provided context, for example when answering questions about a document, policy, or passage \cite{nguyen2024sfrrag}. In other settings, the user expects the model to use retrieved context only as auxiliary evidence and remain skeptical when that context is misleading or conflicts with world knowledge \cite{wang2025conflictingrag,ge2025confact}. A fixed RAG behavior is therefore insufficient: always trusting context may improve contextual faithfulness but can harm factual correctness under misleading retrieval. Conversely, always discounting retrieved context may recover factuality against corrupted evidence but undermine user intent in strict context-following scenarios \cite{fadeeva2025franq}. Practical RAG systems therefore require an intent-aware mechanism that can follow context when the user asks for contextual grounding, while resisting context when the user seeks factual robustness against unreliable retrieval.

To address this challenge, we argue that RAG requires not only retrieval and grounding with external evidence, but also an explicit decoding-time arbitration mechanism that decides when generation should be controlled by retrieved context and when it should rely on parametric memory. In this work, we propose \textbf{Intent-Guided Decoding} (IGD), a decoding-time framework to address both factuality and faithfulness in RAG. In Figure~\ref{fig:igd}, IGD decomposes generation into three conditional branches: a \emph{user branch} conditioned on the original user prompt, a \emph{context branch} that explicitly follows the retrieved context, and a \emph{memory branch} that answers in a closed-book manner. IGD then performs source arbitration at two granularities. First, an \emph{Answer-Level Memory Filter} handles high-confidence cases by routing to a stable memory answer when the retrieved context is likely misleading. Second, for the remaining cases, we propose a novel \emph{Token-Level Correction} mechanism to adjust the user branch decoding distribution only when intervention is needed. Specifically, we leverage an activation gate to detect context-memory conflict, a confidence measurement to decide whether generation should move toward context or memory, and reliability scaling to determine the strength of the intervention. As a result, IGD pulls decoding toward memory when retrieved context is unreliable, while still pulling decoding toward context when the user explicitly requests faithful context following.

Our method IGD introduces a novel decoding-time control layer for RAG, which shifts the focus from retrieving better evidence to arbitrating which knowledge source should steer generation. Unlike retrieval-centric methods that improve what or when to retrieve~\cite{jiang2023flare,asai2024selfrag,yan2024crag}, IGD assumes retrieval has already occurred and addresses downstream conflict between retrieved context and parametric memory. Also, unlike context-faithfulness alignment methods that generally encourage stronger adherence to retrieved evidence~\cite{bi2024contextdpo,zhang2025faithfulrag}, IGD performs bidirectional trust calibration. It adapts the source preference according to user intent. Moreover, unlike prompt-based confidence reasoning or multi-agent evidence aggregation, IGD performs lightweight logit arbitration during generation. Its token-level correction is activated only when the context and memory branches exhibit distributional conflict, enabling local, conservative and intent-aware intervention without retraining the model or replacing the retrieval pipeline.

Our contributions are summarized as follows. First, we formulate the factuality and faithfulness trade-off in RAG as an \emph{intent-conditioned source arbitration} problem, where the appropriate trust policy depends on whether the user requests truth seeking robustness or strict context following. Second, we introduce IGD, a decoding-time framework that combines answer-level memory filtering with conservative token-level correction over user, context, and memory branches. Third, we evaluated and analysis the performance of IGD on three faithful benchmark settings and three factual-conflict benchmark settings. Fourth, we compare IGD against Direct RAG, confidence reasoning baselines such as SCR/RCR, and a multi-agent baseline in strict mode, and further test whether IGD preserves context-following behavior when the user explicitly asks to follow the provided context. Lastly, we release the replication package in~\ref{fn:repo}. 
\section{Related Work}
Retrieval-augmented generation combines parametric language models (LM) with external evidence to improve knowledge intensive generation \cite{lewis2020rag,zhang2025retrieval}. Early and representative RAG systems differ in how they retrieve, encode, and consume evidence. Fusion-in-Decoder conditions generation on multiple retrieved passages and fuses evidence inside the decoder~\cite{izacard2021fid}. Atlas jointly trains a retrieval-augmented model for few-shot knowledge intensive learning~\cite{izacard2023atlas}. In-Context RALM shows that LMs can benefit from retrieved documents inserted directly into the prompt~\cite{ram2023incontext}. Recent work has made retrieval more adaptive, with most remaining retrieval-centric by improving the retriever or training retrieval-aware generators. FLARE retrieves during generation based on predicted future content~\cite{jiang2023flare}, Self-RAG trains models to retrieve and critique the outputs using reflection tokens~\cite{asai2024selfrag}, and CRAG evaluates retrieval quality and corrects low-confidence retrieved evidence~\cite{yan2024crag}.

Other works study hallucination, contextual faithfulness, and knowledge conflict in RAG. RAGTruth provides a corpus to analyze hallucinations in RAG and shows that retrieved evidence does not eliminate unsupported generation~\cite{niu2024ragtruth}. FaithEval evaluates whether models remain faithful under unanswerable, inconsistent, and counterfactual contexts~\cite{ming2024faitheval}. ClashEval studies conflicts between an LLM's internal prior and external evidence, showing that models can adopt incorrect retrieved content even when their internal knowledge is correct~\cite{wu2024clasheval}. Context-DPO improves context faithfulness through preference optimization~\cite{bi2024contextdpo}, while FaithfulRAG models fact-level conflicts between retrieved context and parametric knowledge to improve context-faithful generation~\cite{zhang2025faithfulrag}. Work on situated faithfulness further argues that LLMs should calibrate trust in external context based on both contextual evidence and internal confidence~\cite{huang2024situatedfaithfulness}. Our method shares this motivation of trust calibration, but targets a different control point. Instead of uniformly enforcing stronger adherence to retrieved context, IGD performs intent-aware source arbitration directly in the decoding distribution, allowing the model to favor either context or memory according to the prompt mode.
\section{Method}
\label{sec:method}
Intent-Guided Decoding (IGD) addresses context memory conflict by separating source arbitration into two decision granularities.
Our method first applies an answer-level memory filter, which performs hard replacement only for high-confidence cases, all remaining examples are handled by token-level correction, which continuously adjusts the decoding distribution around the original user prompt. Figure~\ref{fig:igd} illustrates this token-level stage: IGD compares the context and memory branch distributions to activate correction only under meaningful conflict, estimates which branch is more reliable through entropy based confidence and source specific reliability, and then applies a signed logit correction that shifts the user branch distribution toward either the context or memory source.

\subsection{Conditioned Branches}
\label{subsec:method_branches}
At decoding step $t$, let $p_{\mathrm{user},t}(v)$ denote the next token distribution induced by the original user prompt for candidate token $v$. In a standard RAG prompt, this distribution already reflects the interaction between the user instruction, the question, and the retrieved context. IGD augments this original branch with two conditional branches:
\begin{itemize}
    \item $p_{\mathrm{ctx},t}(v)$: a \emph{context-following branch} explicitly instructed to answer according to the supplied context.
    \item $p_{\mathrm{mem},t}(v)$: a \emph{memory branch} instructed to answer in a closed-book manner, independent of the supplied context.
\end{itemize}
The user branch is the unmodified RAG branch: it conditions on the original instruction, question, and retrieved context. In contrast, the context branch is a deliberately source specific branch, we instantiate it with a support snippet selected from the retrieved context, rather than the full context, so that $p_{\mathrm{ctx},t}$ represents the local evidence most relevant to the question. The support snippet is selected using a lightweight IDF-based lexical localizer over context segments, following the classical term-specificity weighting idea of inverse document frequency~\cite{sparckjones1972term}. 

The context branch therefore estimates what the model would generate under a strict context-following prompt, while the memory branch estimates what the model's parametric knowledge supports when the retrieved context is removed. To reduce prompt sensitivity in closed-book decoding, we implement the memory branch as an ensemble of $M$ prompt variants, following the common use of ensembling to stabilize model predictions \cite{dietterich2000ensemble}. In our implementation, $M=3$, corresponding to a base memory prompt, an explicitly closed-book prompt, and a best-guess memory prompt:
\begin{equation}
    p_{\text{mem},t}(v)
    =
    \frac{1}{M}
    \sum_{i=1}^{M}
    p_{\text{mem},t}^{(i)}(v).
    \label{eq:method_mem_ensemble}
\end{equation}

\subsection{Answer-Level Memory Filter}
\label{subsec:method_answer_filter}
The first stage of IGD performs answer-level filtering before decoding, particularly for high-confidence cases in which the memory branch assigns substantially higher likelihood to the memory answer than to the context answer, while the original user branch does not prefer the context answer over the memory answer. Let $\hat{y}_{\mathrm{ctx}}$ denote a short preview answer generated by the context branch, and let $\hat{y}_{\mathrm{mem}}$ denote the memory preview selected from the memory ensemble. For a candidate answer $y=(y_1,\ldots,y_L)$ and branch $b$, we define the length-normalized answer likelihood, and for the memory ensemble we average this likelihood across memory views:

\begin{equation}
\begin{aligned}
    \ell_b(y)
    &=
    \frac{1}{L}
    \sum_{\ell=1}^{L}
    \log p_b(y_\ell \mid y_{<\ell}), \\
    \ell_{\text{mem-ens}}(y)
    &=
    \frac{1}{M}
    \sum_{i=1}^{M}
    \ell_{\text{mem}}^{(i)}(y).
\end{aligned}
\label{eq:method_answer_likelihoods}
\end{equation}
The filter evaluates the memory candidate against the context candidate under two distributions:
\begin{align}
    m_M
    &=
    \ell_{\text{mem-ens}}(\hat{y}_{\text{mem}})
    -
    \ell_{\text{mem-ens}}(\hat{y}_{\text{ctx}}),
    \label{eq:method_margin_mem}
    \\
    m_U
    &=
    \ell_{\text{user}}(\hat{y}_{\text{mem}})
    -
    \ell_{\text{user}}(\hat{y}_{\text{ctx}}).
    \label{eq:method_margin_user}
\end{align}
Here, $m_M$ measures whether the memory branch itself strongly prefers the memory candidate, while $m_U$ checks whether the same candidate remains compatible with the full user prompt. This second condition prevents the filter from overriding the user instruction with a memory answer that the original prompt distribution strongly disfavors. Since this stage performs hard replacement, IGD uses a conservative dominance test rather than an averaged routing score:
\begin{equation}
    D_{\mathrm{mem}}
    =
    \min\!\left(
        m_M-\log\rho_M,\;
        m_U-\log\rho_U
    \right),
    \label{eq:method_memory_dominance}
\end{equation}
where $\rho_M$ and $\rho_U$ are likelihood-ratio thresholds. Since $m_M$ and $m_U$ are differences of length-normalized log likelihoods, $\exp(m_M)$ and $\exp(m_U)$ can be interpreted as length-normalized candidate-level likelihood ratios. IGD routes directly to the memory preview only if
\begin{equation}
\begin{aligned}
    \operatorname{valid}_{\mathrm{mem}} &= 1, \\
    \hat y_{\mathrm{ctx}} &\not\equiv \hat y_{\mathrm{mem}}, \\
    A_{\mathrm{mem}} &\ge 0.67, \\
    D_{\mathrm{mem}} &\ge 0 .
\end{aligned}
\label{eq:IGD_answer_route}
\end{equation}
When these conditions hold, IGD sets $\hat y_{\mathrm{final}}=\hat y_{\mathrm{mem}}$; otherwise, decoding falls back to token-level correction. We use $\rho_M=3.0$ and $\rho_U=1.0$ by default, requiring the memory branch to prefer the memory candidate by at least a factor of three while the user branch must not prefer the context candidate.

\subsection{Token-Level Correction}
\label{subsec:method_token_correction}
IGD performs token-level source arbitration around the original user distribution, the goal is not to replace the distribution, but to gently adjust it when the context and memory branches disagree. This follows the general intuition of decoding time control, where generation is steered by modifying next token scores rather than updating model parameters \cite{krause2020gedi,liu2021dexperts,li2022contrastive}.
\begin{equation}
    p_{\text{final},t}(v)
    \propto
    p_{\text{user},t}(v)
    \left(
        \frac{p_{\text{ctx},t}(v)}
             {p_{\text{mem},t}(v)}
    \right)^{\lambda_t}
    \label{eq:method_main_prob}
\end{equation}
The scalar $\lambda_t$ controls both the direction and magnitude of the intervention, positive values favor the context branch, negative values favor the memory branch, and values near zero leave decoding close to the original user distribution. IGD computes $\lambda_t$ in three steps: \emph{activation}, \emph{direction}, and \emph{reliability scaling}.

\subsubsection{Activation}
\label{subsubsec:method_activation}
IGD should intervene only when context and memory provide meaningfully different next token distribution. We therefore measure distributional conflict at step $t$ between the context and memory branches using Jensen-Shannon divergence \cite{lin1991divergence}:
\begin{equation}
    \delta_t
    =
    \operatorname{JSD}
    \bigl(
        p_{\text{ctx},t}
        \,\| \,
        p_{\text{mem},t}
    \bigr)
    \label{eq:method_delta}
\end{equation}
For efficiency, $\operatorname{JSD}(\cdot\|\cdot)$ is computed over a renormalized top-$K$ token set rather than the full vocabulary we use $K=16$. The conflict score is converted into a soft activation gate:
\begin{equation}
    a_t
    =
    \left[
        \operatorname{clip}
        \left(
            \frac{\delta_t-\tau_{\text{low}}}
                 {\tau_{\text{high}}-\tau_{\text{low}}},
            0,1
        \right)
    \right]^{\gamma}
    \label{eq:method_activation}
\end{equation}
with $\tau_{\mathrm{low}}=0.10$, $\tau_{\mathrm{high}}=0.35$, and $\gamma=2.0$. This gate suppresses correction when the two branches already agree and gradually activates intervention as source conflict increases.

\subsubsection{Confidence Direction}
\label{subsubsec:method_direction}
Conditioned on activation, IGD determines which source should be favored. We combine two signals: an instruction mode and branch level confidence, the instruction encodes the user's intended trust policy or hard context-follow policy, while the branch confidence estimates how concentrated each branch's next token distribution is. We use entropy-based confidence, which is a standard proxy for predictive uncertainty and calibration behavior in neural models, for each branch $b\in\{\mathrm{ctx},\mathrm{mem}\}$, we define:
\begin{equation}
    r_{b,t}
    =
    1
    -
    \frac{
        H(\operatorname{topK}(p_{b,t}))
    }{
        \log K
    }
    \label{eq:method_entropy_conf}
\end{equation}
where $H(\cdot)$ is the entropy of the renormalized top-$K$ distribution, and lower entropy corresponds to higher confidence. Let $d_{\text{mode}}\in(0,1)$ denote the prior probability of trusting context under the current instruction mode, we use $d_{\text{strict}}=0.9$, and $d_{\text{truth}}=0.3$.
The resulting context preference is:
\begin{equation}
    \pi_{\text{ctx},t}
    =
    \sigma
    \left(
        \operatorname{logit}(d_{\text{mode}})
        +
        \log(r_{\text{ctx},t})
        -
        \log(r_{\text{mem},t})
    \right)
    \label{eq:method_pi_ctx}
\end{equation}
and the corresponding signed base coefficient is:
\begin{equation}
    \lambda_{\text{base},t}
    =
    \lambda_{\max}\,
    a_t\,
    \bigl(2\pi_{\text{ctx},t}-1\bigr)
    \label{eq:method_lambda_base}
\end{equation}
This parameterization induces a intent consistent intervention direction: strict context following prompts start with a strong context prior, truth seeking prompts start with a memory prior, and the final direction can still change when one branch is substantially more confident than the other.

\subsubsection{Reliability Scaling}
\label{subsubsec:method_strength}
The base coefficient determines the direction of intervention, IGD then rescales its magnitude by the reliability of the source being favored. This prevents confident but poorly supported context from exerting excessive influence, and similarly prevents unstable memory predictions from overriding context.

\newcommand{\Supp}{\operatorname{Supp}}
\newcommand{\Qual}{\operatorname{Qual}}
\paragraph{Context reliability}
When $\lambda_{\mathrm{base},t}>0$, IGD favors the context branch and estimates whether the selected snippet genuinely supports the context preview. Let $\hat{y}_{\mathrm{ctx}}$ be the context preview and $x_{\mathrm{ctx}}$ be the selected support snippet, we define:
\begin{equation}
\begin{aligned}
    s_{\mathrm{ctx}}
    =
    0.7 \cdot
    \Supp(\hat{y}_{\mathrm{ctx}}, x_{\mathrm{ctx}})
    +
    0.3 \cdot
    \Qual(x_{\mathrm{ctx}})
\end{aligned}
\label{eq:method_context_support}
\end{equation}
where $\operatorname{Supp}(\cdot,\cdot)$ measures local textual support for the preview answer and $\operatorname{Qual}(\cdot)$ captures the evidential quality of selected snippet. Let $\operatorname{valid}_{\mathrm{ctx}}\in\{0,1\}$ indicate whether the context preview is a valid short-form answer, the context reliability is:
\begin{equation}
    q_{\text{ctx}}
    =
    q_{\text{ctx}}^{\min}
    +
    (1-q_{\text{ctx}}^{\min})
    \operatorname{valid}_{\text{ctx}}
    s_{\text{ctx}}
    \label{eq:method_q_ctx}
\end{equation}

\paragraph{Memory reliability}
When $\lambda_{\mathrm{base},t}<0$, IGD favors the memory branch and scales the intervention by the stability of the closed-book memory previews.
\begin{equation}
    q_{\text{mem}}
    =
    q_{\text{mem}}^{\min}
    +
    (1-q_{\text{mem}}^{\min})
    \operatorname{valid}_{\text{mem}}
    A_{\text{mem}}
    \label{eq:method_q_mem}
\end{equation}
We use $q_{\mathrm{ctx}}^{\min}=0.35$ and $q_{\mathrm{mem}}^{\min}=0.55$. Here, $A_{\text{mem}}$ is computed as the average pairwise answer consistency among the $M$ closed-book memory previews:
\begin{equation}
\begin{aligned}
    A_{\mathrm{mem}}
    &=
    \frac{2}{M(M-1)}
    \sum_{1 \le i < j \le M} \\
    &\quad
    \operatorname{AnsCons}
    \Big(
        \hat{y}_{\mathrm{mem}}^{(i)},
        \hat{y}_{\mathrm{mem}}^{(j)}
    \Big)
\end{aligned}
\label{eq:method_a_mem}
\end{equation}
where $\operatorname{AnsCons}(\cdot,\cdot)\in[0,1]$ measures whether two short-form memory previews are equivalent answers.

\paragraph{Final coefficient}
The final token-level coefficient is:
\begin{equation}
\lambda_t
=
\begin{cases}
    \lambda_{\text{base},t}\,q_{\text{ctx}},
    & \lambda_{\text{base},t}>0,\\[4pt]
    \lambda_{\text{base},t}\,q_{\text{mem}},
    & \lambda_{\text{base},t}<0,\\[4pt]
    0,
    & \lambda_{\text{base},t}=0.
\end{cases}
\label{eq:method_lambda_final}
\end{equation}
Thus, IGD intervenes only under source conflict, chooses the direction according to user intent and branch confidence, and scales the update by the reliability of the favored source.
\footnote{\label{fn:repo}\href{https://anonymous.4open.science/r/When-Context-Misleads-Intent-Guided-Decoding-for-Robust-Retrieval-Augmented-Generation-06D7/}{Github Repository}}
\section{Experiment Setup}
\label{sec:experiment_setup}

\subsection{Evaluation Design}
We evaluate IGD in retrieval-augmented short form answering questions. Each example consists of a question, a document-tagged context, and a prompt mode. We consider two prompt modes that represent different user intents. In \textsc{strict} mode, the model is instructed to \textbf{answer strictly according to the provided context}. In \textsc{truth} mode, the model is instructed to \textbf{treat the context as potentially noisy or misleading and to prioritize the factually correct answer when context and world knowledge conflict}. The two modes use the same question and context, only the user instruction changes. Direct RAG and IGD are evaluated under identical prompts, to isolate the effect of source arbitration during the decoding phase.

We organize the evaluation into two groups. The \emph{\textbf{faithful}} group contains standard QA settings where the context supports the gold answer, to test whether IGD preserves the benefits of retrieval. The \emph{\textbf{factual-conflict}} group contains examples where the context answer conflicts with the world answer, to test whether IGD follows the context in \textsc{strict} mode while recovering the world-knowledge answer in \textsc{truth} mode.

\subsection{Benchmarks}
We evaluate on six QA benchmarks, grouped into three faithful QA benchmarks and three factual-conflict QA benchmarks, each contains 500 samples. The faithful group includes KILT-NQ, TriviaQA, and SQuAD. KILT-NQ is based on Natural Questions and the KILT Wikipedia corpus \cite{kwiatkowski2019naturalquestions,petroni2021kilt}; TriviaQA contains trivia style open-domain questions with answer aliases and Wikipedia/web evidence \cite{joshi2017triviaqa}; and SQuAD is a paragraph level reading comprehension dataset built from Wikipedia \cite{rajpurkar2016squad}. For KILT-NQ and TriviaQA, we construct multi-document contexts by selecting an answer bearing passage and adding distractors; for SQuAD, we use the original paragraph as support and add distractor paragraphs from other examples. The factual-conflict group includes ConflictBank, NQ-Swap, and CounterFact. ConflictBank contains Wikidata-derived knowledge conflicts with generated misleading evidence \cite{su2024conflictbank}; NQ-Swap is constructed from Natural Questions/MRQA by replacing the original answer in the context with a substituted answer \cite{kwiatkowski2019naturalquestions,fisch2019mrqa}; and CounterFact is adapted from the factual association editing benchmark in ROME \cite{meng2022rome}. In these factual-conflict benchmarks, the original answer is treated as the world answer, while the misleading or substituted answer supported by the context is treated as the context answer. We apply alias cleaning and semantic filtering to remove ambiguous cases where the world and context answers are equivalent.

\subsection{Models and Baselines}
We evaluate five instruction-tuned LLMs: Qwen3-32B \cite{yang2025qwen3}, Qwen2.5-14B-Instruct \cite{yang2024qwen25}, Llama-3-8B-Instruct \cite{grattafiori2024llama3}, Mistral-7B-Instruct-v0.3 \cite{jiang2023mistral}, and Phi-4 14B \cite{abdin2024phi4}. These models cover different families and scales, allowing us to test whether IGD generalizes across models with different parametric knowledge and instruction-following behavior. Experiments were conducted on local GPU servers equipped with NVIDIA RTX A6000 GPU and two NVIDIA GeForce RTX 5090 GPUs.

We compare IGD with six baselines. \textbf{Closed-book Q-only} receives only the question and no context, it is a diagnostic reference for parametric knowledge rather than a context using RAG method. \textbf{Direct RAG} receives the same question, context, and prompt-mode instruction as IGD, but generates directly from the original prompt without source arbitration. We also include confidence reasoning baselines from situated faithfulness work \cite{huang2024situatedfaithfulness}: \textbf{ExplicitSCR}, which asks the model to explicitly reason about whether to trust the context or its internal knowledge, and three rule-based confidence reasoning variants, \textbf{RCR-InternalEval}, \textbf{RCR-ContextEval}, and \textbf{RCR-InternalConf}, which extract confidence or evaluation signals for internal and context-based answers and then select a final answer according to predefined rules. In addition, we compare with \textbf{MADAM-RAG}, a multi-agent RAG framework designed to aggregate and resolve conflicting retrieved evidence \cite{wang2025conflictingrag}.

\subsection{Evaluation Metrics}
We evaluate the answer accuracy using normalized alias matching, following standard QA evaluation practice \cite{rajpurkar2016squad,joshi2017triviaqa}. For faithful benchmarks, predictions are always matched against gold answers. For factual-conflict benchmarks, the target depends on the prompt mode: \textsc{strict} mode reports \emph{context accuracy}, while \textsc{truth} mode reports \emph{world accuracy}.

Our main aggregate metric is the \textbf{Intent-Aligned Score} (IA), defined as the macro-average over the faithful and factual-conflict benchmark groups:
\begin{equation}
\mathrm{IA}^{m} =
\frac{
\sum_{d \in \mathcal{D}_{\mathrm{faith}}} \mathrm{Acc}^{\mathrm{gold}}_{d}
+
\sum_{d \in \mathcal{D}_{\mathrm{conf}}} \mathrm{Acc}^{m}_{d}
}{
|\mathcal{D}_{\mathrm{faith}}| + |\mathcal{D}_{\mathrm{conf}}|
}.
\end{equation}
Here, $m \in \{\textsc{strict}, \textsc{truth}\}$. For $d \in \mathcal{D}_{\mathrm{conf}}$, $\mathrm{Acc}^{m}_{d}$ denotes context accuracy in \textsc{strict} mode and world accuracy in \textsc{truth} mode, IA therefore measures whether the model behavior matches the user's intended trust policy rather than rewarding a single fixed preference for either context or memory.

\section{Results}
\label{sec:results}

\definecolor{CmpGainGreen}{HTML}{1B8A3A}
\definecolor{CmpLossRed}{HTML}{B3261E}
\definecolor{CmpTieGray}{HTML}{666666}
\definecolor{CmpLightGray}{HTML}{F3F3F3}

\newlength{\cmpIAW}
\newlength{\cmpBenchValW}
\newlength{\cmpGapW}
\newlength{\cmpDeltaW}

\setlength{\cmpIAW}{2.85em}
\setlength{\cmpBenchValW}{2.70em}
\setlength{\cmpGapW}{0.42em}
\setlength{\cmpDeltaW}{2.20em}

\newcommand{\CmpIA}[1]{%
  \makebox[\cmpIAW][r]{#1}%
  \hspace{\cmpGapW}%
  \makebox[\cmpDeltaW][l]{}%
}

\newcommand{\CmpIABest}[1]{%
  \makebox[\cmpIAW][r]{\textbf{#1}}%
  \hspace{\cmpGapW}%
  \makebox[\cmpDeltaW][l]{}%
}

\newcommand{\CmpIASecond}[1]{%
  \makebox[\cmpIAW][r]{\underline{#1}}%
  \hspace{\cmpGapW}%
  \makebox[\cmpDeltaW][l]{}%
}

\newcommand{\CmpIAIGDUpBest}[2]{%
  \makebox[\cmpIAW][r]{\textbf{#1}}%
  \hspace{\cmpGapW}%
  \makebox[\cmpDeltaW][l]{\scriptsize\textcolor{CmpGainGreen}{$\uparrow$#2}}%
}

\newcommand{\CmpIAIGDUpSecond}[2]{%
  \makebox[\cmpIAW][r]{\underline{#1}}%
  \hspace{\cmpGapW}%
  \makebox[\cmpDeltaW][l]{\scriptsize\textcolor{CmpGainGreen}{$\uparrow$#2}}%
}

\newcommand{\CmpIAIGDUp}[2]{%
  \makebox[\cmpIAW][r]{#1}%
  \hspace{\cmpGapW}%
  \makebox[\cmpDeltaW][l]{\scriptsize\textcolor{CmpGainGreen}{$\uparrow$#2}}%
}

\newcommand{\CmpIADash}{%
  \makebox[\cmpIAW][r]{--}%
  \hspace{\cmpGapW}%
  \makebox[\cmpDeltaW][l]{}%
}

\newcommand{\CmpBench}[1]{%
  \makebox[\cmpBenchValW][r]{#1}%
  \hspace{\cmpGapW}%
  \makebox[\cmpDeltaW][l]{}%
}

\newcommand{\CmpBenchBest}[1]{%
  \makebox[\cmpBenchValW][r]{\textbf{#1}}%
  \hspace{\cmpGapW}%
  \makebox[\cmpDeltaW][l]{}%
}

\newcommand{\CmpBenchSecond}[1]{%
  \makebox[\cmpBenchValW][r]{\underline{#1}}%
  \hspace{\cmpGapW}%
  \makebox[\cmpDeltaW][l]{}%
}

\newcommand{\CmpBenchIGDUp}[2]{%
  \makebox[\cmpBenchValW][r]{#1}%
  \hspace{\cmpGapW}%
  \makebox[\cmpDeltaW][l]{\scriptsize\textcolor{CmpGainGreen}{$\uparrow$#2}}%
}

\newcommand{\CmpBenchIGDUpBest}[2]{%
  \makebox[\cmpBenchValW][r]{\textbf{#1}}%
  \hspace{\cmpGapW}%
  \makebox[\cmpDeltaW][l]{\scriptsize\textcolor{CmpGainGreen}{$\uparrow$#2}}%
}

\newcommand{\CmpBenchIGDUpSecond}[2]{%
  \makebox[\cmpBenchValW][r]{\underline{#1}}%
  \hspace{\cmpGapW}%
  \makebox[\cmpDeltaW][l]{\scriptsize\textcolor{CmpGainGreen}{$\uparrow$#2}}%
}

\newcommand{\CmpBenchIGDDown}[2]{%
  \makebox[\cmpBenchValW][r]{#1}%
  \hspace{\cmpGapW}%
  \makebox[\cmpDeltaW][l]{\scriptsize\textcolor{CmpLossRed}{$\downarrow$#2}}%
}

\newcommand{\CmpBenchIGDDownSecond}[2]{%
  \makebox[\cmpBenchValW][r]{\underline{#1}}%
  \hspace{\cmpGapW}%
  \makebox[\cmpDeltaW][l]{\scriptsize\textcolor{CmpLossRed}{$\downarrow$#2}}%
}

\newcommand{\CmpBenchDash}{%
  \makebox[\cmpBenchValW][r]{--}%
  \hspace{\cmpGapW}%
  \makebox[\cmpDeltaW][l]{}%
}

\newcommand{\CmpExternalMissingRows}{%
& ExplicitSCR
& \CmpIADash
& \CmpBenchDash
& \CmpBenchDash
& \CmpBenchDash
& \CmpBenchDash
& \CmpBenchDash
& \CmpBenchDash \\

& RCR-InternalEval
& \CmpIADash
& \CmpBenchDash
& \CmpBenchDash
& \CmpBenchDash
& \CmpBenchDash
& \CmpBenchDash
& \CmpBenchDash \\

& RCR-ContextEval
& \CmpIADash
& \CmpBenchDash
& \CmpBenchDash
& \CmpBenchDash
& \CmpBenchDash
& \CmpBenchDash
& \CmpBenchDash \\

& RCR-InternalConf
& \CmpIADash
& \CmpBenchDash
& \CmpBenchDash
& \CmpBenchDash
& \CmpBenchDash
& \CmpBenchDash
& \CmpBenchDash \\

& MADAM-RAG
& \CmpIADash
& \CmpBenchDash
& \CmpBenchDash
& \CmpBenchDash
& \CmpBenchDash
& \CmpBenchDash
& \CmpBenchDash \\
}

\begin{table*}[t]
\centering
\footnotesize
\caption{
Truth-mode comparison across three faithful benchmarks and three factual-conflict benchmarks.
IA is the macro-average over the six benchmark columns.
Bold marks the best context-using method in each column, and underline marks the second-best.
IGD also reports deltas against Direct RAG under the same model and benchmark.
}
\label{tab:IGD_truth_six_benchmarks}
\setlength{\tabcolsep}{2.2pt}
\renewcommand{\arraystretch}{1.08}

\begin{threeparttable}
\resizebox{\textwidth}{!}{
\begin{tabular}{llc|ccc|ccc}
\toprule
\multirow{2}{*}{Model} &
\multirow{2}{*}{Method} &
\multirow{2}{*}{IA} &
\multicolumn{3}{c|}{Faithful benchmarks} &
\multicolumn{3}{c}{Factual-conflict benchmarks} \\
\cmidrule(lr){4-6}\cmidrule(lr){7-9}
& & &
KILT-NQ &
TriviaQA &
SQuAD &
ConflictBank &
NQ-Swap &
CounterFact \\
\midrule

\rowcolor{CmpLightGray}

\textbf{Qwen3-32B}
& Closed-book Q-only$^{\dagger}$
& \CmpIA{56.8}
& \CmpBench{35.4}
& \CmpBench{62.0}
& \CmpBench{32.4}
& \CmpBench{56.4}
& \CmpBench{70.0}
& \CmpBench{84.6} \\

& Direct RAG
& \CmpIA{51.0}
& \CmpBenchBest{74.8}
& \CmpBenchBest{88.8}
& \CmpBenchBest{90.4}
& \CmpBench{15.0}
& \CmpBench{27.0}
& \CmpBench{10.0} \\

& ExplicitSCR
& \CmpIA{12.9}
& \CmpBench{16.0}
& \CmpBench{18.4}
& \CmpBench{15.8}
& \CmpBench{4.0}
& \CmpBench{3.0}
& \CmpBench{20.2} \\

& RCR-InternalEval
& \CmpIA{62.1}
& \CmpBench{58.2}
& \CmpBench{74.6}
& \CmpBench{47.8}
& \CmpBenchSecond{47.8}
& \CmpBench{70.5}
& \CmpBench{73.6} \\

& RCR-ContextEval
& \CmpIA{53.6}
& \CmpBench{68.8}
& \CmpBench{80.0}
& \CmpBench{83.2}
& \CmpBench{21.0}
& \CmpBench{37.0}
& \CmpBench{31.4} \\

& RCR-InternalConf
& \CmpIA{59.9}
& \CmpBench{50.0}
& \CmpBench{69.0}
& \CmpBench{41.8}
& \CmpBench{46.6}
& \CmpBenchBest{77.5}
& \CmpBenchSecond{74.8} \\

& MADAM-RAG
& \CmpIASecond{65.0}
& \CmpBench{62.6}
& \CmpBench{82.4}
& \CmpBench{66.2}
& \CmpBench{44.0}
& \CmpBench{61.0}
& \CmpBench{73.8} \\

& IGD
& \CmpIAIGDUpBest{74.0}{23.0}
& \CmpBenchIGDDownSecond{72.2}{2.6}
& \CmpBenchIGDDownSecond{84.8}{4.0}
& \CmpBenchIGDDownSecond{89.8}{0.6}
& \CmpBenchIGDUpBest{50.2}{35.2}
& \CmpBenchIGDUp{71.5}{44.5}
& \CmpBenchIGDUpBest{75.4}{65.4} \\

\midrule

\rowcolor{CmpLightGray}

\textbf{Qwen2.5-14B}
& Closed-book Q-only$^{\dagger}$
& \CmpIA{63.3}
& \CmpBench{37.2}
& \CmpBench{65.6}
& \CmpBench{31.2}
& \CmpBench{61.2}
& \CmpBench{91.5}
& \CmpBench{92.8} \\

& Direct RAG
& \CmpIA{58.1}
& \CmpBenchBest{74.2}
& \CmpBenchBest{90.2}
& \CmpBenchBest{94.2}
& \CmpBench{12.0}
& \CmpBench{42.0}
& \CmpBench{36.0} \\

& ExplicitSCR
& \CmpIA{10.8}
& \CmpBench{13.0}
& \CmpBench{16.4}
& \CmpBench{14.2}
& \CmpBench{4.8}
& \CmpBench{9.5}
& \CmpBench{6.6} \\

& RCR-InternalEval
& \CmpIA{63.5}
& \CmpBench{64.8}
& \CmpBench{80.6}
& \CmpBench{65.0}
& \CmpBench{39.2}
& \CmpBench{57.0}
& \CmpBench{74.6} \\

& RCR-ContextEval
& \CmpIA{59.6}
& \CmpBench{57.8}
& \CmpBench{76.4}
& \CmpBench{69.8}
& \CmpBench{26.4}
& \CmpBench{57.5}
& \CmpBench{69.8} \\

& RCR-InternalConf
& \CmpIASecond{65.0}
& \CmpBench{54.8}
& \CmpBench{75.6}
& \CmpBench{49.2}
& \CmpBenchSecond{55.0}
& \CmpBench{67.0}
& \CmpBenchSecond{88.2} \\

& MADAM-RAG
& \CmpIA{59.7}
& \CmpBench{57.4}
& \CmpBench{71.8}
& \CmpBench{61.0}
& \CmpBench{38.2}
& \CmpBenchSecond{63.0}
& \CmpBench{66.6} \\

& IGD
& \CmpIAIGDUpBest{77.6}{19.5}
& \CmpBenchIGDDownSecond{70.8}{3.4}
& \CmpBenchIGDDownSecond{85.6}{4.6}
& \CmpBenchIGDDownSecond{87.8}{6.4}
& \CmpBenchIGDUpBest{55.4}{43.4}
& \CmpBenchIGDUpBest{77.0}{35.0}
& \CmpBenchIGDUpBest{89.4}{53.4} \\

\midrule

\rowcolor{CmpLightGray}

\textbf{Llama-3-8B}
& Closed-book Q-only$^{\dagger}$
& \CmpIA{55.7}
& \CmpBench{37.8}
& \CmpBench{69.8}
& \CmpBench{26.8}
& \CmpBench{55.6}
& \CmpBench{68.5}
& \CmpBench{75.8} \\

& Direct RAG
& \CmpIA{59.6}
& \CmpBenchBest{73.2}
& \CmpBenchBest{88.4}
& \CmpBenchBest{89.4}
& \CmpBench{27.6}
& \CmpBench{49.0}
& \CmpBench{30.2} \\

& ExplicitSCR
& \CmpIA{11.2}
& \CmpBench{9.8}
& \CmpBench{8.8}
& \CmpBench{8.8}
& \CmpBench{2.4}
& \CmpBench{16.0}
& \CmpBench{21.2} \\

& RCR-InternalEval
& \CmpIASecond{59.7}
& \CmpBench{59.8}
& \CmpBench{81.0}
& \CmpBench{58.8}
& \CmpBench{41.0}
& \CmpBench{55.0}
& \CmpBench{62.8} \\

& RCR-ContextEval
& \CmpIA{53.3}
& \CmpBench{42.6}
& \CmpBench{68.6}
& \CmpBench{33.2}
& \CmpBenchSecond{42.2}
& \CmpBenchSecond{61.5}
& \CmpBenchBest{71.6} \\

& RCR-InternalConf
& \CmpIA{52.1}
& \CmpBench{45.0}
& \CmpBench{75.2}
& \CmpBench{40.6}
& \CmpBench{32.2}
& \CmpBench{56.0}
& \CmpBench{63.8} \\

& MADAM-RAG
& \CmpIA{45.9}
& \CmpBench{51.6}
& \CmpBench{64.6}
& \CmpBench{60.4}
& \CmpBench{28.0}
& \CmpBench{40.0}
& \CmpBench{30.6} \\

& IGD
& \CmpIAIGDUpBest{69.3}{9.6}
& \CmpBenchIGDDownSecond{68.4}{4.8}
& \CmpBenchIGDDownSecond{85.8}{2.6}
& \CmpBenchIGDDownSecond{85.0}{4.4}
& \CmpBenchIGDUpBest{46.8}{19.2}
& \CmpBenchIGDUpBest{62.0}{13.0}
& \CmpBenchIGDUpSecond{67.6}{37.4} \\

\midrule

\rowcolor{CmpLightGray}

\textbf{Mistral-7B}
& Closed-book Q-only$^{\dagger}$
& \CmpIA{51.8}
& \CmpBench{39.2}
& \CmpBench{66.2}
& \CmpBench{24.2}
& \CmpBench{39.4}
& \CmpBench{68.5}
& \CmpBench{73.0} \\

& Direct RAG
& \CmpIA{42.7}
& \CmpBenchBest{61.4}
& \CmpBenchBest{77.4}
& \CmpBenchBest{77.0}
& \CmpBench{11.4}
& \CmpBench{17.5}
& \CmpBench{11.4} \\

& ExplicitSCR
& \CmpIA{26.6}
& \CmpBench{23.2}
& \CmpBench{32.0}
& \CmpBench{22.6}
& \CmpBench{17.6}
& \CmpBench{21.0}
& \CmpBench{43.2} \\

& RCR-InternalEval
& \CmpIA{45.7}
& \CmpBench{50.0}
& \CmpBench{72.8}
& \CmpBench{41.4}
& \CmpBench{26.2}
& \CmpBenchSecond{44.0}
& \CmpBench{40.0} \\

& RCR-ContextEval
& \CmpIA{47.0}
& \CmpBench{57.0}
& \CmpBench{76.4}
& \CmpBench{72.8}
& \CmpBench{10.6}
& \CmpBench{22.5}
& \CmpBench{42.4} \\

& RCR-InternalConf
& \CmpIASecond{50.1}
& \CmpBench{49.6}
& \CmpBench{74.2}
& \CmpBench{38.2}
& \CmpBench{36.6}
& \CmpBench{40.5}
& \CmpBenchBest{61.4} \\

& MADAM-RAG
& \CmpIA{44.4}
& \CmpBench{50.6}
& \CmpBench{59.6}
& \CmpBench{58.6}
& \CmpBench{19.2}
& \CmpBenchBest{52.5}
& \CmpBench{25.6} \\

& IGD
& \CmpIAIGDUpBest{54.2}{11.5}
& \CmpBenchIGDDownSecond{58.4}{3.0}
& \CmpBenchIGDDownSecond{76.0}{1.4}
& \CmpBenchIGDDownSecond{73.0}{4.0}
& \CmpBenchIGDUpBest{37.0}{25.6}
& \CmpBenchIGDUp{43.0}{25.5}
& \CmpBenchIGDUpSecond{47.8}{36.4} \\

\midrule

\rowcolor{CmpLightGray}

\textbf{Phi-4 14B}
& Closed-book Q-only$^{\dagger}$
& \CmpIA{51.8}
& \CmpBench{32.6}
& \CmpBench{66.6}
& \CmpBench{28.2}
& \CmpBench{41.8}
& \CmpBench{60.0}
& \CmpBench{81.4} \\

& Direct RAG
& \CmpIA{56.9}
& \CmpBenchBest{71.0}
& \CmpBenchBest{87.2}
& \CmpBenchBest{85.8}
& \CmpBench{15.8}
& \CmpBench{35.5}
& \CmpBench{46.2} \\

& ExplicitSCR
& \CmpIA{6.1}
& \CmpBench{11.8}
& \CmpBench{14.6}
& \CmpBench{3.0}
& \CmpBench{2.6}
& \CmpBench{2.0}
& \CmpBench{2.8} \\

& RCR-InternalEval
& \CmpIA{55.8}
& \CmpBench{58.8}
& \CmpBench{77.6}
& \CmpBench{49.2}
& \CmpBench{30.8}
& \CmpBench{47.5}
& \CmpBenchSecond{70.8} \\

& RCR-ContextEval
& \CmpIA{57.5}
& \CmpBenchSecond{65.8}
& \CmpBench{81.8}
& \CmpBench{66.8}
& \CmpBench{21.6}
& \CmpBench{45.5}
& \CmpBench{63.4} \\

& RCR-InternalConf
& \CmpIA{55.7}
& \CmpBench{54.0}
& \CmpBench{76.2}
& \CmpBench{45.8}
& \CmpBenchSecond{32.8}
& \CmpBench{53.0}
& \CmpBench{72.4} \\

& MADAM-RAG
& \CmpIASecond{61.1}
& \CmpBench{61.6}
& \CmpBench{79.8}
& \CmpBench{59.0}
& \CmpBench{32.2}
& \CmpBenchSecond{57.0}
& \CmpBenchBest{77.2} \\

& IGD
& \CmpIAIGDUpBest{67.1}{10.2}
& \CmpBenchIGDDown{63.8}{7.2}
& \CmpBenchIGDDownSecond{85.8}{1.4}
& \CmpBenchIGDDownSecond{82.2}{3.6}
& \CmpBenchIGDUpBest{34.8}{19.0}
& \CmpBenchIGDUpBest{59.0}{23.5}
& \CmpBenchIGDUp{77.0}{30.8} \\

\bottomrule
\end{tabular}
}
\end{threeparttable}
\end{table*}

\begin{figure*}
    \centering
    \includegraphics[width=\linewidth]{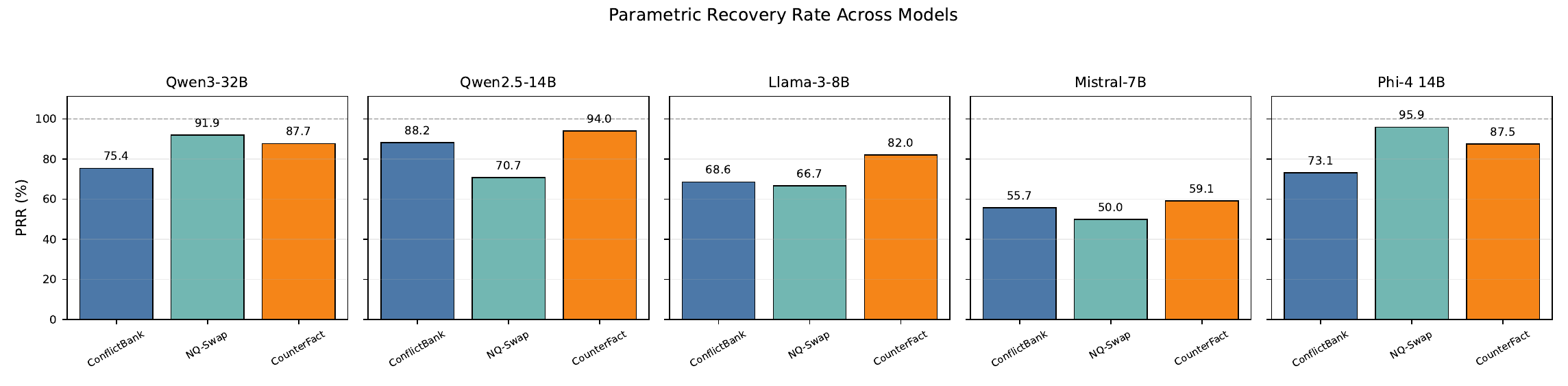}
    \caption{
    Parametric recovery rate (PRR) of IGD on factual-conflict benchmarks, the dashed line denotes full recovery relative to the closed-book reference.
    }
    \label{fig:recover}
\end{figure*}

\subsection{Closed-Book Performance Reveals the Recoverable Parametric Signal}
\label{subsec:closed_book_diagnostic}
We first evaluate each model in the closed-book Q-only setting, where the model receives only the question without any retrieved context. 
It is important to interpret IGD because factual recovery from a misleading context is only possible when the underlying model can access the correct answer from its parametric memory, if the closed-book model cannot recover the world answer, the upper bound of any memory oriented correction is naturally limited  \cite{mallen2023whennot,kandpal2023longtail}.

The closed-book results in Table~\ref{tab:IGD_truth_six_benchmarks} show that the difficulty of the six benchmarks differs substantially. Among the faithful benchmarks, closed-book accuracy is generally low on KILT-NQ and SQuAD, indicating that these datasets rely heavily on the provided context \cite{kwiatkowski2019naturalquestions,petroni2021kilt,rajpurkar2016squad}. Across the five models, KILT-NQ closed-book accuracy remains around the mid 30\% range, and SQuAD is even lower, mostly around 25-32\%. This is expected: KILT-NQ is derived from open-domain queries grounded in Wikipedia evidence, while SQuAD questions are written against specific paragraphs and often require access to the original passage. In contrast, TriviaQA is substantially easier in the closed-book setting, with all models reaching above 60\%, suggesting that many trivia-style answers are already encoded in the models' parameters.

For the factual-conflict benchmarks, closed-book performance is much higher overall, especially on CounterFact and NQ-Swap. CounterFact is the easiest factual-conflict benchmark under closed-book evaluation: all models achieve between 73.0\% and 92.8\%, suggesting that many subject relation facts remain accessible from parametric memory. NQ-Swap is also highly recoverable for several models, with Qwen2.5-14B reaching 91.5\% and Phi-4 reaching 60.0\%. ConflictBank is more challenging, with closed-book accuracy ranging from 39.4\% to 61.2\%. Overall, Qwen2.5-14B-Instruct shows the strongest closed-book IA score, while Qwen3-32B also exhibits strong recoverability on the factual-conflict benchmarks. These diagnostics suggest that IGD is evaluated in a meaningful regime: the factual-conflict examples are difficult for Direct RAG because the context is misleading, but many of their correct answers are still recoverable from model memory.

\subsection{Truth Prompting Alone Does Not Calibrate Context Trust}
\label{subsec:direct_rag_failure}

Table~\ref{tab:IGD_truth_six_benchmarks} shows that explicit truth seeking instructions alone are insufficient to prevent RAG models from being misled by incorrect context. In truth mode, the prompt explicitly instructs the model to prioritize the correct answer and to treat the provided context with skepticism when it may be misleading \cite{wu2024clasheval,huang2024situatedfaithfulness}. Nevertheless, Direct RAG performs poorly on factual-conflict benchmarks across all models, this failure is most visible when comparing Direct RAG with the closed-book diagnostic. For example, Qwen2.5-14B answers NQ-Swap correctly 91.5\% of the time in the closed-book setting, but drops to 42.0\% once the misleading context is provided. The same model reaches 92.8\% closed-book accuracy on CounterFact, but Direct RAG falls to 36.0\%. 

The additional baselines further clarify that this problem is not fully solved by answer-level reasoning or confidence-based source selection. ExplicitSCR and RCR variants are designed to improve situated faithfulness by reasoning over, or extracting confidence from, internal and context-based answers \cite{huang2024situatedfaithfulness}, while MADAM-RAG uses multi-agent aggregation to handle conflicting retrieved evidence \cite{wang2025conflictingrag}. These methods often improve over Direct RAG on factual-conflict benchmarks, confirming that explicit source comparison is useful. However, their improvements are uneven: some variants recover more world answers but substantially reduce faithful QA accuracy, while others preserve context grounded performance but remain weak under misleading context. In contrast, IGD achieves the best IA score for every evaluated backbone in Table~\ref{tab:IGD_truth_six_benchmarks}. This suggests that the key difficulty is not only detecting whether context or memory is correct at the answer level, but also controlling how strongly each source should influence decoding. the desired behavior is not to simply suppress context, but to preserve context grounded accuracy on faithful examples while resisting misleading evidence in factual conflict cases.


\definecolor{GainGreen}{HTML}{1B8A3A}
\definecolor{LossRed}{HTML}{B3261E}
\definecolor{TieGray}{HTML}{666666}
\definecolor{LightGray}{HTML}{F3F3F3}

\newcommand{\numstrong}[1]{%
  \scalebox{0.94}{\textbf{#1}}%
}

\newlength{\iaw}
\newlength{\benchvalw}
\newlength{\benchgapw}
\newlength{\benchdeltaw}

\setlength{\iaw}{2.85em}
\setlength{\benchvalw}{2.70em}
\setlength{\benchgapw}{0.42em}
\setlength{\benchdeltaw}{2.20em}

\newcommand{\IAcell}[1]{%
  \makebox[\iaw][r]{#1}%
}
\newcommand{\IAcellbf}[1]{%
  \makebox[\iaw][r]{\numstrong{#1}}%
}

\newcommand{\Bench}[1]{%
  \makebox[\benchvalw][r]{#1}%
  \hspace{\benchgapw}%
  \makebox[\benchdeltaw][l]{}%
}
\newcommand{\Benchbf}[1]{%
  \makebox[\benchvalw][r]{\numstrong{#1}}%
  \hspace{\benchgapw}%
  \makebox[\benchdeltaw][l]{}%
}

\newcommand{\BenchUp}[2]{%
  \makebox[\benchvalw][r]{\numstrong{#1}}%
  \hspace{\benchgapw}%
  \makebox[\benchdeltaw][l]{\scriptsize\textcolor{GainGreen}{$\uparrow$#2}}%
}
\newcommand{\BenchDown}[2]{%
  \makebox[\benchvalw][r]{#1}%
  \hspace{\benchgapw}%
  \makebox[\benchdeltaw][l]{\scriptsize\textcolor{LossRed}{$\downarrow$#2}}%
}
\newcommand{\BenchTie}[2]{%
  \makebox[\benchvalw][r]{\numstrong{#1}}%
  \hspace{\benchgapw}%
  \makebox[\benchdeltaw][l]{\scriptsize\textcolor{TieGray}{$\leftrightarrow$#2}}%
}
\begin{table*}[t]
\footnotesize
\centering
\caption{Strict context following prompt. Factual-conflict benchmarks are evaluated by context accuracy, testing whether IGD preserves user intent when the user explicitly asks to follow the provided context.}
\label{tab:IGD_strict_six_benchmarks}
\setlength{\tabcolsep}{2.2pt}
\renewcommand{\arraystretch}{1.08}
\begin{threeparttable}
\resizebox{\textwidth}{!}{
\begin{tabular}{llc|ccc|ccc}
\toprule
\multirow{2}{*}{Model} &
\multirow{2}{*}{Method} &
\multirow{2}{*}{IA} &
\multicolumn{3}{c|}{Faithful benchmarks} &
\multicolumn{3}{c}{Factual-conflict benchmarks} \\
\cmidrule(lr){4-6}\cmidrule(lr){7-9}
& & &
KILT-NQ &
TriviaQA &
SQuAD &
ConflictBank &
NQ-Swap &
CounterFact \\
\midrule

\textbf{Qwen3-32B}
& Direct RAG
& \IAcell{83.0}
& \Benchbf{72.6}
& \Benchbf{89.4}
& \Benchbf{91.0}
& \Bench{83.2}
& \Bench{63.0}
& \Bench{99.0} \\
& IGD
& \IAcellbf{84.3}
& \BenchDown{69.4}{ 3.2}
& \BenchDown{86.8}{ 2.6}
& \BenchDown{90.0}{ 1.0}
& \BenchUp{87.2}{ 4.0}
& \BenchUp{72.5}{ 9.5}
& \BenchUp{99.6}{ 0.6} \\

\midrule

\textbf{Qwen2.5-14B}
& Direct RAG
& \IAcell{84.2}
& \Bench{74.6}
& \Benchbf{89.2}
& \Benchbf{94.0}
& \Bench{84.4}
& \Bench{63.0}
& \Benchbf{100.0} \\
& IGD
& \IAcellbf{88.1}
& \BenchUp{76.4}{ 1.8}
& \BenchDown{88.6}{ 0.6}
& \BenchDown{93.6}{ 0.4}
& \BenchUp{89.2}{ 4.8}
& \BenchUp{81.0}{ 18.0}
& \BenchTie{100.0}{ 0.0} \\

\midrule

\textbf{Llama-3-8B}
& Direct RAG
& \IAcell{80.8}
& \Benchbf{73.0}
& \Benchbf{88.0}
& \Benchbf{88.4}
& \Bench{85.4}
& \Bench{50.0}
& \Bench{99.8} \\
& IGD
& \IAcellbf{82.0}
& \BenchDown{71.0}{ 2.0}
& \BenchDown{86.4}{ 1.6}
& \BenchDown{87.6}{ 0.8}
& \BenchUp{88.0}{ 2.6}
& \BenchUp{59.0}{ 9.0}
& \BenchUp{100.0}{ 0.2} \\

\midrule

\textbf{Mistral-7B}
& Direct RAG
& \IAcell{77.1}
& \Benchbf{64.0}
& \Benchbf{79.6}
& \Bench{80.8}
& \Bench{82.6}
& \Bench{59.5}
& \Bench{96.0} \\
& IGD
& \IAcellbf{77.6}
& \BenchDown{60.6}{ 3.4}
& \BenchDown{79.2}{ 0.4}
& \BenchUp{82.0}{ 1.2}
& \BenchUp{84.4}{ 1.8}
& \BenchUp{62.5}{ 3.0}
& \BenchUp{97.0}{ 1.0} \\

\midrule

\textbf{Phi-4 14B}
& Direct RAG
& \IAcell{81.3}
& \Benchbf{73.2}
& \Benchbf{84.0}
& \Bench{86.6}
& \Bench{82.4}
& \Bench{63.0}
& \Bench{98.4} \\
& IGD
& \IAcellbf{82.5}
& \BenchDown{71.6}{ 1.6}
& \BenchDown{83.2}{ 0.8}
& \BenchUp{87.0}{ 0.4}
& \BenchUp{85.6}{ 3.2}
& \BenchUp{68.5}{ 5.5}
& \BenchUp{99.2}{ 0.8} \\

\bottomrule
\end{tabular}
}
\end{threeparttable}
\end{table*}

\subsection{IGD Balances Factuality and Faithfulness}
\label{subsec:igd_balances_tradeoff}

\paragraph{Truth mode: recovering world answers under misleading context.}
Table~\ref{tab:IGD_truth_six_benchmarks} shows that IGD substantially improves truth mode factual recovery while largely preserving faithful QA accuracy. Compared with Direct RAG, IGD improves the IA score for every model, with gains of 23.0 points for Qwen3-32B, 19.5 for Qwen2.5-14B, 9.6 for Llama-3-8B, 11.5 for Mistral-7B, and 10.2 for Phi-4. These gains are driven primarily by large improvements on factual-conflict benchmarks, where IGD consistently shifts generation away from misleading context and toward recoverable world knowledge.

The comparison with external baselines provides a more informative view of this trade-off. Confidence-reasoning and multi-agent baselines can be strong on individual factual-conflict columns, but they often sacrifice accuracy on faithful benchmarks. For instance, RCR-style methods may recover more world-knowledge answers on a particular conflict set, but their faithful accuracy drops sharply because they rely on a global answer-level source decision. Similarly, MADAM-RAG improves some conflict cases, but its aggregation strategy does not consistently preserve original context-grounded behavior.



The factual gains are large and systematic. Across all models, IGD improves every factual-conflict benchmark over Direct RAG, with the largest gain reaching 65.4 percentage points on Qwen3-32B CounterFact. Importantly, these gains do not come from discarding retrieval altogether. On faithful benchmarks, IGD remains close to Direct RAG, with moderate drops that are much smaller than the factual-conflict gains. This pattern suggests that token-level correction is more conservative than answer-level rerouting alone: it can weaken misleading context when necessary, while still preserving useful retrieved evidence when the context is correct.

To better quantify how much of the recoverable parametric signal is restored, Figure~\ref{fig:recover} reports the parametric recovery rate (PRR), defined as the fraction of the gap between Direct RAG and closed-book Q-only that IGD recovers on factual-conflict benchmarks:
\begin{equation}
\mathrm{PRR}
=
\frac{
\mathrm{Acc}_{\mathrm{IGD}} - \mathrm{Acc}_{\mathrm{Direct}}
}{
\mathrm{Acc}_{\mathrm{Closed}} - \mathrm{Acc}_{\mathrm{Direct}}
}.
\end{equation}

The PRR results show that IGD recovers a large portion of the available parametric knowledge for stronger models. For example, Qwen3-32B recovers most of performance gap on NQ-Swap and CounterFact, whereas Mistral-7B exhibits weaker recovery. This indicates that IGD's factual correction is bounded by both the model's accessible parametric knowledge and the reliability of the memory branch. The pattern in Figure~\ref{fig:recover} therefore reinforces the interpretation that IGD is not hallucinating new facts, but recovering parametric answers that are otherwise suppressed by misleading context.

\paragraph{Strict mode: preserving context-following intent}
Table~\ref{tab:IGD_strict_six_benchmarks} evaluates the complementary setting: the user explicitly asks the model to follow the provided context. In this mode, factual-conflict benchmarks are evaluated by context accuracy rather than world accuracy, because the desired behavior is to follow the supplied evidence even when it conflicts with world knowledge. Since the external baselines in Table~\ref{tab:IGD_truth_six_benchmarks} are designed primarily for truth-seeking or correctness-oriented source selection, we keep the strict-mode comparison focused on Direct RAG and IGD.

The results show that IGD preserves, and in many cases improves, strict context-following behavior. These strict prompt's mode results are central to the claim that IGD balances factuality and faithfulness rather than optimizing only one side of the trade-off \cite{fadeeva2025franq}. A method that simply discounts retrieved context would improve truth mode factual accuracy but damage strict context following \cite{bi2024contextdpo,zhang2025faithfulrag}. In contrast, IGD uses the prompt mode to determine the direction of intervention: under truth-seeking prompts it can pull decoding toward memory, while under strict context-following prompts it preserves or strengthens context guided decoding. 

\section{Factuality and Faithfulness Trade-off}
\newcommand{\severedrop}[1]{\cellcolor{red!16}\textbf{#1}}
\newcommand{\largedrop}[1]{\cellcolor{red!9}#1}
\newcommand{\gaincell}[1]{\cellcolor{green!8}#1}

\begin{table}[t]
\centering
\caption{
Core ablations of IGD on Qwen2.5-14B.
Values are macro-averaged percentages over the three faithful benchmarks and the three factual-conflict benchmarks.
Shaded cells mark notable changes relative to IGD full: light red indicates drops of roughly 3--10 percentage points, darker red indicates drops larger than 10 points, and light green indicates improvements over IGD full.
}
\label{tab:igd_core_ablation}

\begingroup
\scriptsize
\setlength{\tabcolsep}{5.0pt}
\renewcommand{\arraystretch}{1.12}

\begin{threeparttable}
\begin{adjustbox}{width=0.95\linewidth,center}
\begin{tabular}{l l cc cc}
\toprule
\multirow{2}{*}{Variant} &
\multirow{2}{*}{Removed / changed component} &
\multicolumn{2}{c}{Strict mode} &
\multicolumn{2}{c}{Truth mode} \\
\cmidrule(lr){3-4}
\cmidrule(lr){5-6}
&
&
\makecell[c]{Faithful\\Gold} &
\makecell[c]{Factual\\Ctx} &
\makecell[c]{Faithful\\Gold} &
\makecell[c]{Factual\\World} \\
\midrule
\rowcolor{gray!10}
IGD full
& none
& \textbf{86.2} & \textbf{90.1}
& 81.4 & \textbf{73.7} \\
\midrule

w/o Answer Filter
& answer-level memory filter
& 85.8 & \largedrop{86.0}
& 81.3 & \severedrop{63.0} \\

w/o Token Correction
& token-level correction
& \gaincell{86.9} & \largedrop{83.1}
& \gaincell{82.7} & \severedrop{52.8} \\

w/o Activation Gate
& JSD conflict gate
& 85.9 & \largedrop{86.2}
& \largedrop{77.5} & \largedrop{70.5} \\

w/o Mode Prior
& instruction-mode prior
& \gaincell{86.6} & \largedrop{83.4}
& \gaincell{82.7} & \severedrop{63.1} \\

w/o Reliability Gate
& $q_{\mathrm{ctx}}/q_{\mathrm{mem}}$ scaling
& 85.0 & \largedrop{85.5}
& \largedrop{76.1} & \largedrop{68.6} \\

Single Memory View
& memory ensemble
& 86.1 & \largedrop{85.8}
& \gaincell{85.5} & \severedrop{54.5} \\
\bottomrule
\end{tabular}
\end{adjustbox}
\end{threeparttable}

\endgroup
\end{table}
We further analyze how IGD balances factual recovery and context faithfulness through component ablations and hyperparameter sensitivity. This analysis is motivated by the central tension in RAG, stronger reliance on parametric memory can recover world answers but may undermine strict context following \cite{ming2024faitheval,huang2024situatedfaithfulness}. We conduct ablations on Qwen2.5-14B-Instruct over the same six-benchmark suite used in the main experiments, reporting macro averaged results over the three faithful benchmarks and the three factual-conflict benchmarks. Table~\ref{tab:igd_core_ablation} ablates six core components of IGD: the answer level memory filter, token-level correction, activation gate, instruction-mode prior, reliability scaling, and memory ensemble, we choose these ablations because each component directly affects the trade-off rather than merely changing implementation details.

Table~\ref{tab:igd_core_ablation} shows that IGD's gains arise from coordinated source arbitration rather than a single heuristic. Removing token-level correction causes the largest degradation, reducing truth-mode factual world accuracy from 73.7 to 52.8 and strict-mode factual context accuracy from 90.1 to 83.1. This indicates that token-level correction is not merely a memory side intervention, but the main mechanism for steering: it pulls decoding toward memory under misleading context while preserving context-following behavior when required. The answer-level memory filter provides a complementary high precision route for stable factual-conflict cases; removing it drops factual world accuracy to 63.0 while leaving faithful accuracy nearly unchanged, suggesting that the filter is conservative enough not to disrupt ordinary context-grounded QA. The remaining ablations explain how IGD avoids over-intervention, a single memory view sharply reduces factual recovery to 54.5, showing that memory side correction requires a stable closed-book signal; removing the activation or reliability gate degrades both faithful and factual columns, indicating that intervention must be triggered only under meaningful conflict and scaled by source reliability. Finally, removing the mode prior lowers both truth-mode factual recovery and strict-mode context accuracy, confirming that context and memory conflicts cannot be resolved with a fixed source preference, whereas the full IGD model provides a stable balance between factual recovery and context faithfulness. 
\begin{figure}[t]
    \centering
    \includegraphics[width=\linewidth]{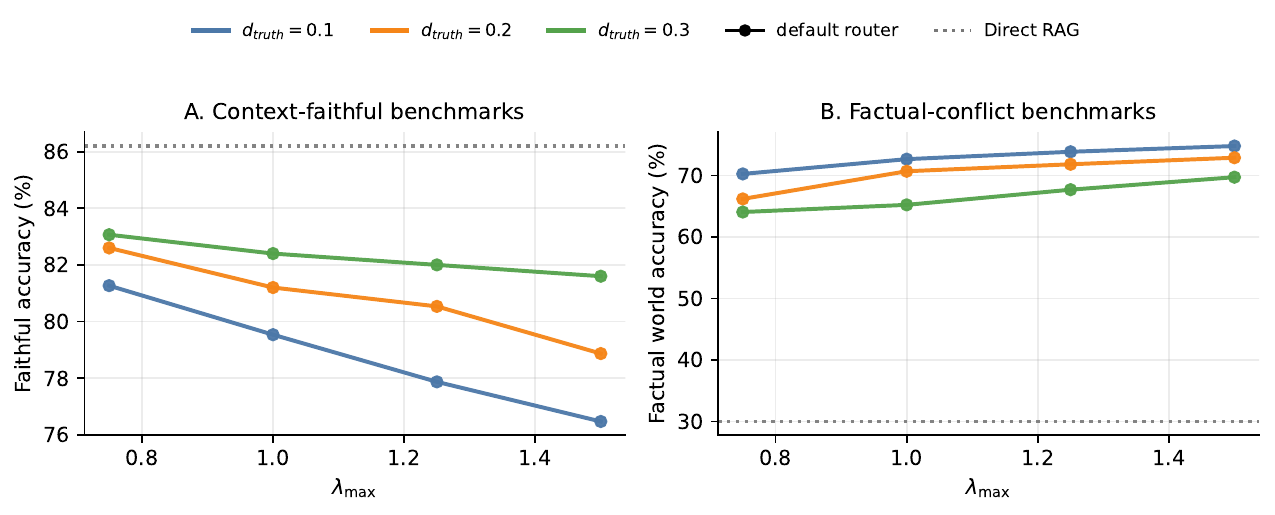}
    \caption{
    Sensitivity of truth mode performance to token-level steering hyperparameters on Qwen2.5-14B-Instruct.
    }
    \label{fig:lambda}
\end{figure}

Figure~\ref{fig:lambda} further exposes the limitation of using fixed global steering hyperparameters. As $\lambda_{\max}$ increases, factual world accuracy on factual-conflict benchmarks generally improves, but faithful accuracy on faithful benchmarks decreases. Similarly, smaller $d_{\mathrm{truth}}$ values make the router more memory biased and improve factual recovery, but they also make the model more likely to deviate from correct retrieved evidence. This creates a clear tuning trade-off: aggressive correction helps resist misleading context, whereas conservative correction better preserves faithful context use, this is a core limitation of the current framework. Future work could reduce this limitation by learning example adaptive intervention strengths, calibrating the router with validation feedback, or estimating source reliability at a finer granularity. 

\begin{figure}
    \centering
    \includegraphics[width=\linewidth]{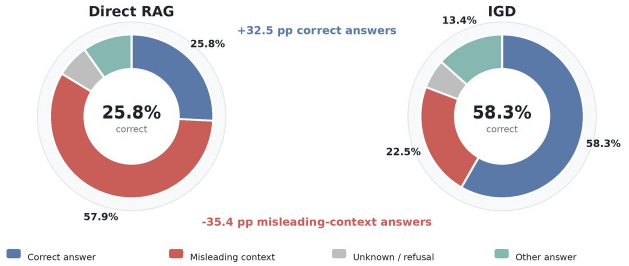}
    \caption{ Answer distribution on factual-conflict benchmarks under truth mode.
    }
    \label{fig:answer_distribution}
\end{figure}
Beyond aggregate accuracy, we further analyze what models output on factual-conflict benchmarks. Main results report whether an answer matches the target, but they do not distinguish whether errors come from copying the misleading context, refusing to answer, or producing an unrelated alternative. We therefore categorize each output into four types: \emph{correct answer}, \emph{misleading-context answer}, \emph{unknown/refusal}, and \emph{other answer}. As shown in Figure~\ref{fig:answer_distribution}, Direct RAG is dominated by misleading context answers: 57.9\% of its outputs follow the incorrect context, while only 25.8\% recover the correct world answer, IGD reverses this pattern, increasing correct answers to 58.3\% and reducing misleading-context answers to 22.5\%. 
At the same time, IGD does not convert all misleading context answers into correct answers, some mass moves into the \emph{other answer} category. We interpret this as a residual uncertainty effect under source conflict: once decoding is pushed away from the misleading context, the model may still fail to land on the gold world answer if the memory branch is unstable, if competing aliases are plausible, or if the evidence conflict is not cleanly resolved. This is consistent with recent work showing that RAG models remain vulnerable under conflicting or noisy evidence and that models often generate plausible incorrect answers rather than abstaining when uncertainty is not explicitly rewarded \cite{ge2025confact,chen2025carerag,kalai2025why}. Thus, IGD substantially mitigates context over-trust, but factual-conflict resolution remains imperfect when memory evidence is ambiguous or insufficiently calibrated.

\begin{figure}
    \centering
    \includegraphics[width=\linewidth]{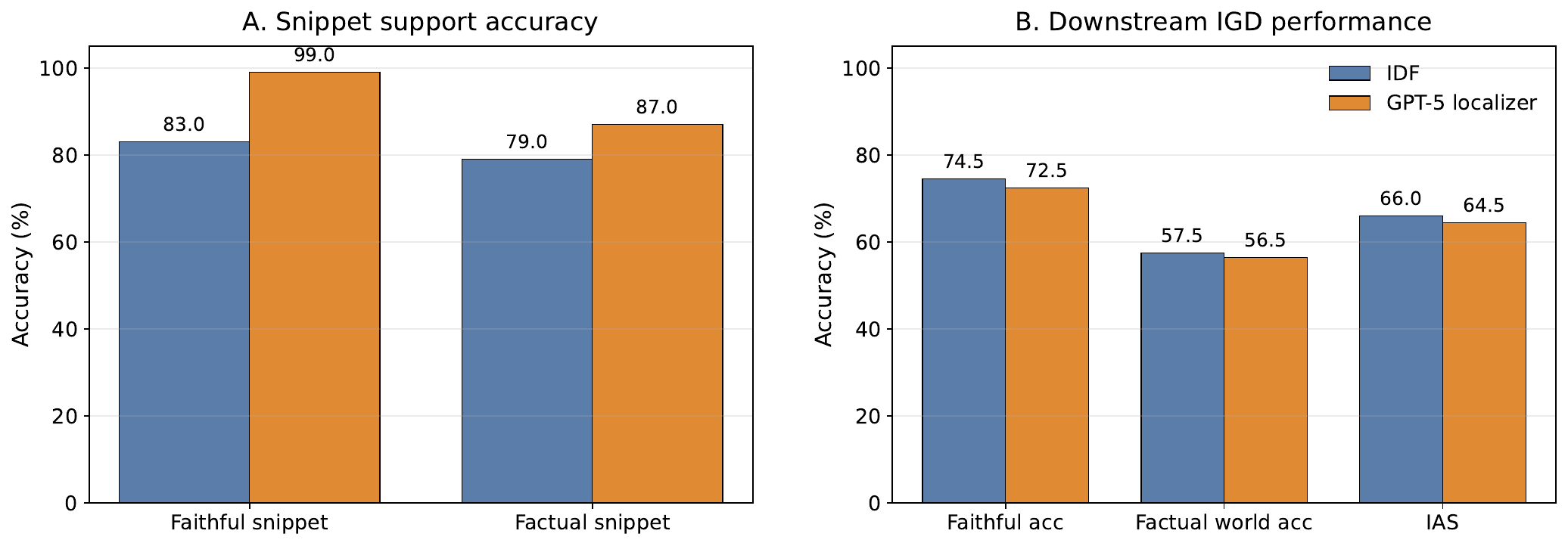}
    \caption{Snippet localization accuracy versus downstream IGD performance.
    }
    \label{fig:localizer}
\end{figure}
Figure~\ref{fig:localizer} highlights a surprising mismatch between snippet localization quality and downstream IGD performance. The GPT-5 localizer substantially improves standalone snippet support accuracy, increasing faithful snippet accuracy from 83.0\% to 99.0\% and factual snippet accuracy from 79.0\% to 87.0\%. Intuitively, one might expect a better support localizer to improve IGD, since IGD uses snippet reliability to scale context-side intervention. However, the downstream results show the opposite: replacing the IDF localizer slightly reduces faithful accuracy, factual world accuracy, and IAS. This suggests that IGD does not only need a snippet that is semantically supportive under human or LLM judgment, it needs a snippet whose evidence is useful for the specific downstream generator and reliability function. The IDF localizer may select shorter, lexically concentrated spans that align better with answer-bearing tokens and IGD's support heuristics, while the GPT-5 localizer may prefer semantically richer snippets that are correct in isolation but less discriminative for logit-level source arbitration \cite{deng2025influence,kim2025ltrr}. The result also aligns with evidence that robust RAG depends on preserving fine-grained textual details, not only high-level semantic support \cite{jin2025sara}. More broadly, LLM-based scoring or localization can introduce its own judgment biases and objective mismatch \cite{li2025scoringbias}.
\section{Conclusion}
\label{sec:conclusion}
We presented Intent-Guided Decoding (IGD), a decoding-time framework for resolving the factuality and faithfulness trade-off in retrieval-augmented generation. Rather than treating retrieved context as either universally authoritative or inherently suspect, IGD performs intent-conditioned source arbitration across the user prompt, retrieved context, and parametric memory. Experiments on faithful and factual-conflict benchmarks show that IGD improves intent-aligned generation across multiple LLMs, while ablations confirm that its gains arise from coordinated routing, conflict activation, reliability scaling, and memory stabilization.

\bibliography{citation}
\bibliographystyle{ieeetr}

\end{document}